\documentclass[sigconf, authorversion]{acmart}  

\AtBeginDocument{%
  }

\setcopyright{acmlicensed}
\copyrightyear{2024}
\acmYear{2024}
\acmDOI{XXXXXXX.XXXXXXX}

\acmConference[xx]{xxx}{xx, 2024}{xx, xx}
\acmISBN{978-1-4503-XXXX-X/18/06}

\renewcommand\footnotetextcopyrightpermission[1]{}  

\usepackage[linesnumbered,ruled,vlined]{algorithm2e}
\usepackage{diagbox}
\usepackage{multirow}
\usepackage{booktabs}
\usepackage{bbding}
\usepackage{amsmath, bm}
\usepackage{graphicx}
\usepackage{subcaption}
\usepackage{arydshln} 
\usepackage{makecell}

\begin{document}

\title{SFedCA: Credit Assignment-Based Active Client Selection Strategy for Spiking Federated Learning}

\author{Qiugang Zhan}
\email{zhanqiugang@std.uestc.edu.cn}
\affiliation{%
  \institution{University of Electronic Science and Technology of China}
  \streetaddress{}
  \city{}
  \state{}
  \country{}
  \postcode{}
}
\affiliation{%
  \institution{Southwestern University of Finance and Economics}
  \streetaddress{}
  \city{Chengdu}
  \state{}
  \country{China}
  \postcode{}
}

\author{Jinbo Cao}
\email{cjinboswufe@163.com}
\affiliation{%
  \institution{Southwestern University of Finance and Economics}
  \streetaddress{}
  \city{Chengdu}
  \state{}
  \country{China}
  \postcode{}
}

\author{Xiurui Xie}
\email{xiexiurui@uestc.edu.cn}
\affiliation{%
  \institution{University of Electronic Science and Technology of China}
  \streetaddress{}
  \city{Chengdu}
  \state{}
  \country{China}
  \postcode{}
}

\author{Malu Zhang}
\email{maluzhang@uestc.edu.cn}
\affiliation{%
  \institution{University of Electronic Science and Technology of China}
  \streetaddress{}
  \city{Chengdu}
  \state{}
  \country{China}
  \postcode{}
}

\author{Huajin Tang}
\email{htang@zju.edu.cn}
\affiliation{%
  \institution{Zhejiang University}
  \streetaddress{}
  \city{Zhejiang}
  \state{}
  \country{China}
  \postcode{}
}

\author{Guisong Liu}
\email{gliu@swufe.edu.cn}
\authornotemark[1]
\affiliation{%
  \institution{Southwestern University of Finance and Economics}
  \streetaddress{}
  \city{Chengdu}
  \state{}
  \country{China}
  \postcode{}
}


\begin{abstract}
Spiking federated learning is an emerging distributed learning paradigm that allows resource-constrained devices to train collaboratively at low power consumption without exchanging local data. 
It takes advantage of both the privacy computation property in federated learning (FL) and the energy efficiency in spiking neural networks (SNN). Thus, it is highly promising to revolutionize the efficient processing of multimedia data.
However, existing spiking federated learning methods employ a random selection approach for client aggregation, assuming unbiased client participation.
This neglect of statistical heterogeneity affects the convergence and accuracy of the global model significantly.
In our work, we propose a credit assignment-based active client selection strategy, the SFedCA, to judiciously aggregate clients that contribute to the global sample distribution balance.
Specifically, the client credits are assigned by the firing intensity state before and after local model training, which reflects the local data distribution difference from the global model.
Comprehensive experiments are conducted on various non-identical and independent distribution (non-IID) scenarios. The experimental results demonstrate that the SFedCA outperforms the existing state-of-the-art spiking federated learning methods, and requires fewer communication rounds. 

\end{abstract}

\ccsdesc[300]{Computing methodologies~Cooperation and coordination}

\keywords{Spiking Neural Network, Federated Learning, Client Selection}


\maketitle

\section{Introduction}

\begin{figure}[h]
  \centering
  \includegraphics[width=\linewidth]{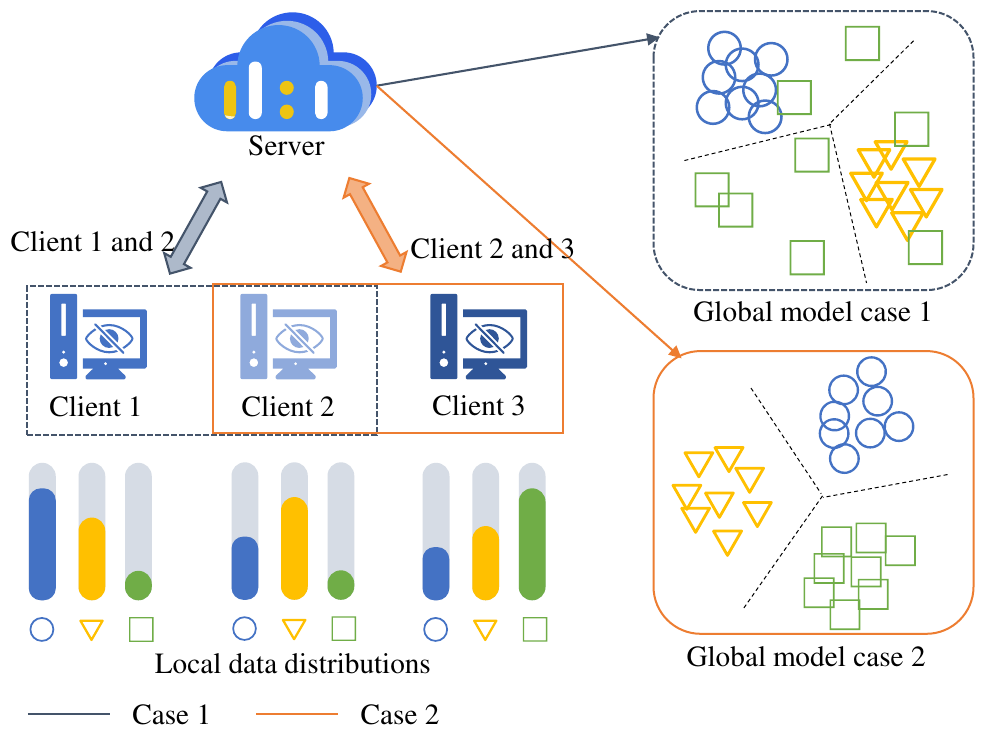}
  \caption{The influence of selecting different clients. In case 1, client 1 and 2 participate in the aggregation; in case 2, client 2 and 3 participate.}
  \label{fig: client selection effect sample}
\end{figure}

With the burst of data growth and evolving multimedia technologies, FL emerges as a transformative approach that promises to address key challenges related to data privacy and data sharing.
In recent years, FL has yielded many achievements based on artificial neural networks (ANNs) in the areas of image and video analytics, speech recognition, and content recommendation, advancing the development of multimedia technologies for privacy and security \cite{lu2023federated, zhang2023affectfal, li2023prototype}.
However, the excessive computational power cost makes traditional ANN-based FL methods difficult to be deployed on resource-constrained edge devices.

Spiking federated learning provides a solution to this problem.
It is an emerging federated learning paradigm that deploys SNN models with low power consumption on clients, and jointly trains them while maintaining data security and privacy \cite{tumpa2023federated, yang2022lead}.
The SNN mimics the dynamical mechanisms of biological neurons and communicates with sparse spiking signals rather than real numbers, thus it has extremely low computational power consumption \cite{zhan2024two}.
The effectiveness of spiking federated learning has been demonstrated in many scenarios such as audio recognition, radar signal recognition, and neuromorphic visual recognition \cite{yang2022lead, venkatesha2021federated, xie2022efficient, skatchkovsky2020federated}.

Despite achieving successful applications in many fields, these spiking federated learning methods only focus on random sampling schemes for client selection, which often damage the accuracy and convergence speed of the global model.
This is because samples of each client follow a non-IID that is known as statistical heterogeneity in real federated systems.
The high degree of statistical heterogeneity makes not all models trained by all clients suitable for the global model, and blindly aggregating all client models tends to reduce the global model efficiency \cite{ma2022state, fu2023client}.
As shown in Figure \ref{fig: client selection effect sample}, when the clients randomly selected (case 1 in Figure \ref{fig: client selection effect sample}) to participate in the aggregation have similar data distributions, the aggregated global model also has a more restricted view.
If the selected clients could cover a wider distribution (case 2 in Figure \ref{fig: client selection effect sample}), the global model also has a stronger classification ability.

To address this issue, researchers have theoretically analyzed the effects of statistical heterogeneity and proposed several active client selection strategies.
These methods select clients to participate in the aggregation based on the differences between the model of clients and the server to accelerate the convergence of the global model on non-IID data \cite{goetz2019active}. 
The latest approaches select clients based on the difference in gradient between the global and local model, and significantly improve model performance on non-IID data \cite{balakrishnan2022diverse, Wang2023FedMoSTC}.
However, these strategies are proposed on ANN-based FL and do not take into account the special training mechanisms of spiking federated learning as well as the proprietary biological neuronal information.

To obtain a spiking federated learning method with resistance to the effects of non-IID, we introduce a credit assignment concept from SNN, which is used to measure the contribution of individual neurons in SNN models \cite{wang2023adaptive}.
We use this concept to describe the contribution of clients to the global model.
Specifically, the assigned credit can be derived from the unique characteristics of spiking neurons associated with training effectiveness.
The firing rates of neurons describe their activity level and excitatory state, providing important insights into the dynamic behavior, information transmission speed, and neural activity patterns of the neural network \cite{li2021towards}.
The firing rate stats also have been demonstrated to be reflective of the training value of the model in SNN \cite{lee2016training, zhan2023bio}. 
Based on the above analysis, we propose a spiking federated method built on the credit assignment-based client selection strategy, called SFedCA.
Our main contributions are summarized as follows.
\begin{itemize}
    \item We present the SFedCA method that realizes effective client selection on spiking federation learning.
    To the best of our knowledge, this is the first attempt to use the active client selection strategy in the spiking federated learning field, which addresses the slow convergence and low accuracy problems brought by non-IID.
    \item We propose a client credit assignment method based on the firing rate difference of spiking neurons. 
    Clients selected by this method can provide a large data distribution scope for the global model updating.
    \item Our method outperforms existing spiking federated learning methods under multiple non-IID distributions, achieving higher accuracy, faster convergence, and more stable performance. 
\end{itemize}

\section{Background and Related Works}
\subsection{Integrate-and-Fire (IF) spiking neuron model}
\begin{figure}[h]
  \centering
  \includegraphics[width=\linewidth]{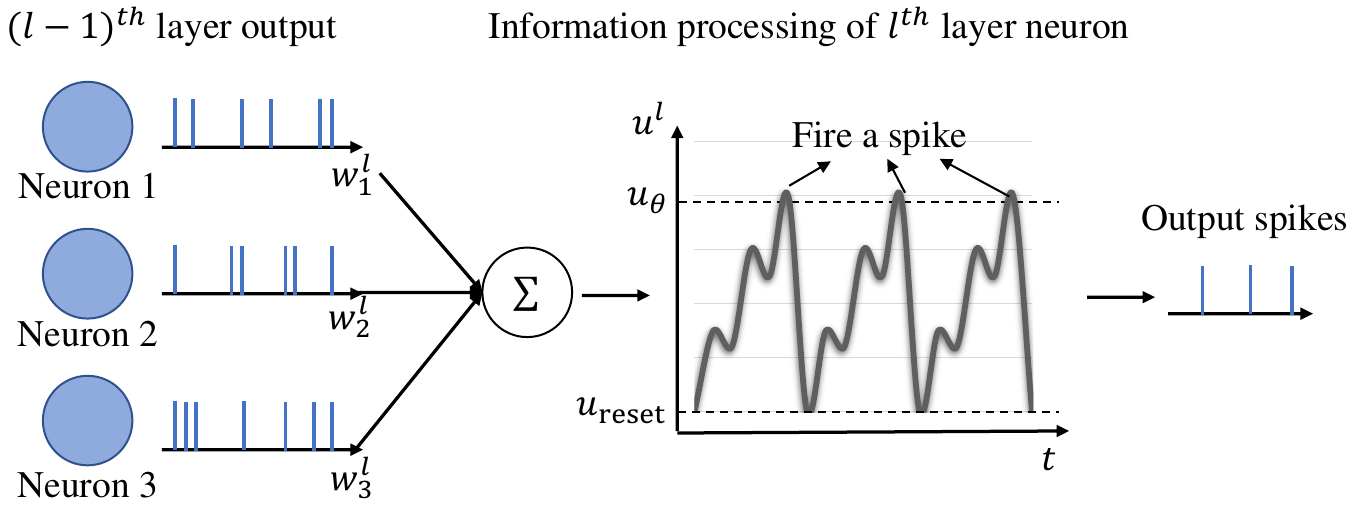}
  \caption{Information processing in IF neuron model. The neuron receives input spikes causing the change of membrane potential, and when it exceeds the threshold, the neuron fires a spike and resets the membrane potential.}
  \label{fig: LIF model}
\end{figure}

A common spiking neural network consists of spiking neurons that simulate biological information processing.
These neurons receive binary spikes as the input and after various membrane potential transformation rules, then these neurons will also emit spike sequences as the output.
Integrate-and-Fire (IF) is one of the most popular neuron models in SNNs \cite{wu2019direct, fang2023spikingjelly, liu2023human}, simulating the changes in membrane potential and current inside neurons in a simple and effective manner. 
The main activities of the IF neuron include charging, discharging, and resetting, represented by the following formulas.
\begin{equation}
    v^{l, t} = u^{l, t-1} + w^l \cdot o^{l-1, t},
    \label{eq: charge}
\end{equation}
\begin{equation}
    o^{l, t}=g\left(v^{l, t}-u_{\theta}\right),
    \label{eq: firing}
\end{equation}
\begin{equation}
    u^{l, t}=v^{l, t}\left(1-o^{l, t}\right)+u_{\text{reset}} \cdot o^{l, t},
    \label{eq: reset}
\end{equation}
where $v^{l, t}$ represents the charging membrane potential under threshold in the $l^{th}$ layer at time $t$; $o^{l, t}$ is the output spikes of the $l^{th}$ layer neurons at time $t$; 
$u^{l, t}$ is the neuronal membrane potential after the spike function; $u_{\theta}$ is the firing threshold; $u_{\text{reset}}$ is the resetting potential; $g(x)$ is the Heaviside function defined as follows:
\begin{equation}
    g(x)=\left\{
            \begin{array}{ll}
                0, & x<0; \\
                1, & x \geq 0.
            \end{array}
          \right .
    \label{eq: step function}
\end{equation}
Figure \ref{fig: LIF model} shows the information processing of the IF neuron.

The SNN models used in this work are constructed by the IF neuron model.


\subsection{Spiking Federated Learning} 
The first spiking federated learning method was proposed by \cite{skatchkovsky2020federated}, which was trained in an online manner and evaluated on two clients.
Current research in this area can be categorized into the following two aspects.
In terms of training methods, \cite{venkatesha2021federated} presented FedSNN which has a more complex network structure and can be applied on a larger client scale by the backpropagation algorithm. 
When the total number of clients is greater than 10, FedSNN selects the clients for aggregation in a randomized way.
Then, Yang et al. proposed a decentralized federated neuromorphic learning method LFNL \cite{yang2022lead}, that dynamically selected clients with high capabilities (e.g., computational and communication capabilities) as leaders aggregating all 4 client models.
In terms of applications, 
reference \cite{xie2022efficient} applied the FL framework with SNNs to distributed traffic sign recognition.
In \cite{zhang2024federated}, spiking FL was used in radar gesture recognition with privacy requirements, and a novel FL approach with distributed parameter pruning was designed to reduce the communication cost of FL.

\begin{figure*}[h]
  \centering
  \includegraphics[width=\linewidth]{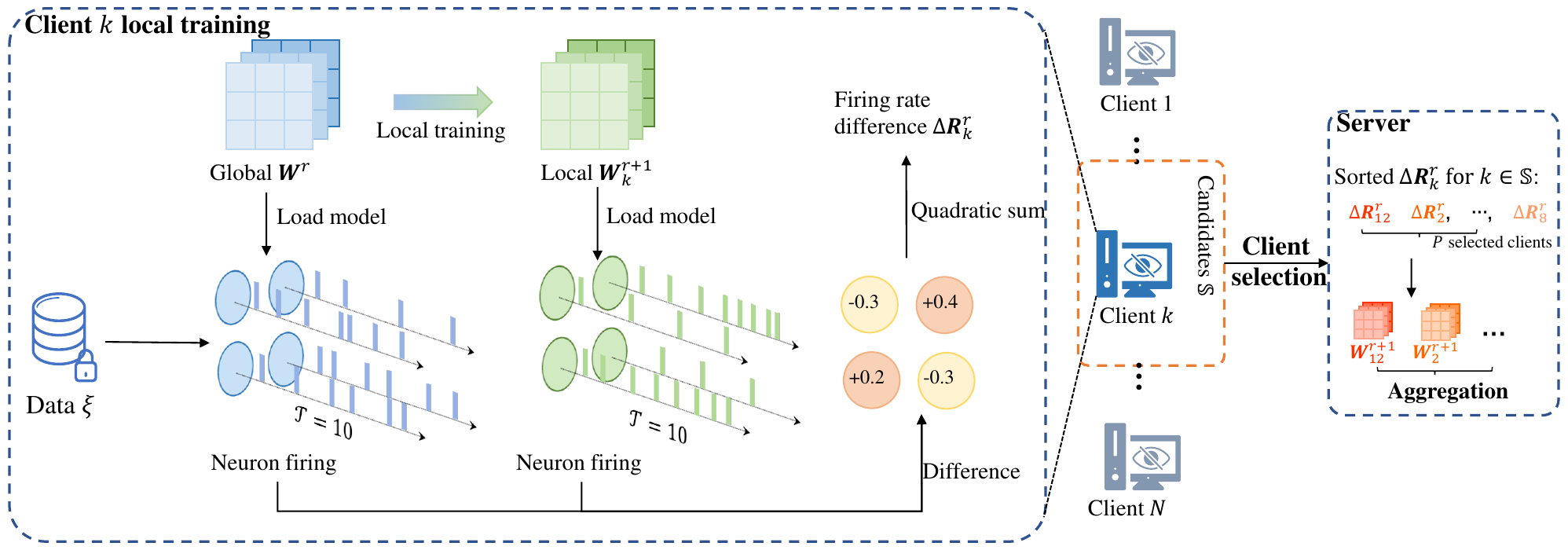}
  \caption{The framework of SFedCA. In the local training process, the client calculates the firing rate difference $\Delta \boldsymbol{R}_k^r$ according to the global model $\boldsymbol{W}^r$ and the updated local model $\boldsymbol{W}_k^{r+1}$. The server selects $P$ clients with higher firing rate differences from client candidates $\mathbb{S}$, and gets the new global model by aggregating the parameters of selected clients.}
  \label{fig: SFedCA}
\end{figure*}

Although current spiking federated learning has yielded considerable research results, these methods either aggregate all clients or randomly select partial clients, ignoring the impact on statistical heterogeneity.
In this paper, we view client selection as a credit assignment problem and use the properties of SNNs to select clients that are ``important'' for training, aiming to improve the accuracy and convergence speed of the global model.

\section{Method}
In this section, We first formally describe the federated learning problem and introduce the local model training process. 
Then, we analyze the relationship between credit allocation and data distribution balancing and present the specific client selection strategy.
\subsection{Problem Formulation}
A common federated learning system consists of $N$ clients and a central server.
Each client, indexed by $k=1, 2, \cdots, N$, maintains a local private dataset $\mathcal{D}_k$.
The typical federated learning objective is:
\begin{equation}
    \min _{\boldsymbol{W}} \mathcal{L}\left(\boldsymbol{W}\right) = \sum_{k=1}^N p_k \mathcal{L}_k\left(\boldsymbol{W}\right),
\end{equation}
where $\mathcal{L}(\boldsymbol{W})$ is the global loss function for model $\boldsymbol{W}$, and the global loss is calculated from the local losses $\mathcal{L}_k(\boldsymbol{W})$ of all clients weighted according to the fraction of data $p_k = \frac{|\mathcal{D}_k|}{\sum_{k=1}^N |\mathcal{D}_k|}$, $|\mathcal{D}_k|$ is the local data scale of client $k$.

At the $r^{th}$ round, the server broadcasts the global model $\boldsymbol{W}^{r}$ to all clients to initiate the local training process. 
Upon receiving the initial model $\boldsymbol{W}^{r}$, each client trains the model on its private dataset for $e$ epochs.
In the federated learning process, the server only takes a subset of clients $\mathbb{P}^r$ with size $|\mathbb{P}^r| = P < N$ for aggregation. 
These clients upload the local trained model $\{\boldsymbol{W}_k^{r+1}\}_{k\in\mathbb{P}^r}$ to the server.
The server aggregates these models (usually by averaging \cite{mcmahan2017communication}) to obtain the global model $\boldsymbol{W}^{r+1}$.
This procedure can be formulated as Eq. \eqref{eq: aggregation}:
\begin{equation}
    \boldsymbol{W}^{r+1} =\boldsymbol{W}^r -  
    \frac{\eta^r}{\sum_{k\in\mathbb{P}^r}|\mathcal{D}_k|}\sum_{k\in\mathbb{P}^r}|\mathcal{D}_k| \nabla \mathcal{L}_k\left(\boldsymbol{W}^r\right),
    \label{eq: aggregation}
\end{equation}
where $\nabla \mathcal{L}_k$ means the gradient of $\mathcal{L}_k$, $\eta^r$ is the learning rate of the $r^{th}$ round.
In this study, we have an easily satisfied assumption that in each round, the local model converges to $\varepsilon$-neighborhood of the optimum on the local data after training for $e$ epochs.

Figure \ref{fig: SFedCA} shows the framework of the proposed SFedCA method.
The client uses local data to train the model and then uploads it to the server. The server aggregates the selected client models and redistributes them to the clients for the next round of training.

\subsection{Local Model Training}
For each client, the local model is implemented by an SNN model. 
The local model training is started by encoding the input data $\xi \in \mathbb{R}^{3\times d \times d}$ into spike trains $x \in \{0, 1\}^{\mathcal{T} \times 3\times d \times d}$, where 
$3\times d \times d$ is the shape of input data and $\mathcal{T}$ is the time window.

For an $L$ layers SNN model, the spike output $o^{l} = \left(o^{l, 1}, o^{l, 2}, \cdots,\notag \right. \\ \left. o^{l, \mathcal{T}}\right)^\mathbf{T}$ of each layer is calculated by Eq. \eqref{eq: charge}, \eqref{eq: firing} and \eqref{eq: reset}, and $\mathbf{T}$ means transpose. 
We use $o^{l, t}(\xi, \boldsymbol{W})$ to represent the spike output of the $l^{th}$ layer at the $t^{th}$ time step, where $\boldsymbol{W} = \left(w^1, w^2, \cdots, w^L\right)^\mathbf{T}$.

The local loss function $\mathcal{L}$ is defined by the cross-entropy loss between the predicted classification probability of each classify neuron $z^{L}(\xi; \boldsymbol{W})$ and the ground truth labels $y$, as shown below:
\begin{equation}
    \mathcal{L}\left(\xi, y; \boldsymbol{W}\right) = - \sum_{c=1}^{C}y[c] \log\left(z^{L}(\xi; \boldsymbol{W})[c]\right),
\end{equation}
where $[c]$ means the component of the vector in the $c^{th}$ category and $C$ is the number of categories.
The predicted classification probability $z^{L}(\xi; \boldsymbol{W})$ is calculated by the firing rate of the last layer neurons as follows:
\begin{equation}
    z^{L}(\xi; \boldsymbol{W})[c] = 
    \frac{\exp \left( \sum_{t=1}^{\mathcal{T}} o^{L, t}(\xi; \boldsymbol{W})[c] \right)}{\sum_{c=1}^{C} \exp \left(\sum_{t=1}^{\mathcal{T}} o^{L, t}(\xi; \boldsymbol{W})[c]\right)}.
\end{equation}

During the backpropagation of the error $\mathcal{L}$, the gradient of the loss with respect to the model weights $\boldsymbol{W}$ is computed according to Eq. \eqref{eq: charge} - \eqref{eq: reset} as follows:

\begin{equation}
    \frac{\partial \mathcal{L}}{\partial \boldsymbol{W}}=\frac{\partial \mathcal{L}}{\partial z}\frac{\partial z}{\partial o}\frac{\partial o}{\partial v}\frac{\partial v}{\partial \boldsymbol{W}}.
    \label{eq: BP}
\end{equation}
The $1^{st}$, $2^{nd}$, and $4^{th}$ terms of Eq. \eqref{eq: BP} can be derived by the chain rule, while the original function of the $3^{rd}$ term i.e. Eq.\eqref{eq: step function} is an undifferentiable function.
Following reference \cite{fang2023spikingjelly}, we take the gradient of the arc tangent function as the surrogate gradient of Eq. \eqref{eq: step function}, which is shown as follows:
\begin{equation}
    \frac{\partial o}{\partial v} = \frac{\alpha}{2 \left(1 + \left( \frac{\pi}{2} \alpha \cdot o\right)\right)},
\end{equation}
where $\alpha$ is a hyperparameter that controls the shape of the surrogate gradient and we set it to 2.0.

\subsection{Credit Assignment and Distribution Balance} \label{sec: Firing Rate and Distribution Balance}
Under the non-IID data distribution cases, clients should be assigned higher credit that can add new distributional information to the global model from the previous round.
The firing rate of the SNN model provides an effective idea to measure the distribution difference without directly comparing the data.
For close data distributions, SNN neurons will exhibit similar spike firing patterns \cite{zhan2023bio}, so models trained on dissimilar distributions will show large differences in firing rates.

To explore the relationship between the credit assignment and distribution balance, we analyze different aggregation cases across different distributions.
Table. \ref{tab: toy experiment} shows an example of MNIST (using categories 0, 1 and 2) with three clients.
In round 1, we aggregate clients 1 and 2, and the accuracy of the global model on the test set is 66.44\%.
Based on this global model, we calculate the firing rate difference before and after local training (2 epochs) on each client.
Client 3 has the largest difference in data distribution used with the global model that aggregates clients 1 and 2, and also has the largest firing rate difference 9.63\%.
The data distributions across clients 1 and 2 exhibit a pronounced imbalance. 
Therefore, when client 3—possessing a distinct data distribution—is omitted from the aggregation in round 2, it results in the lowest test accuracy 66.44\%.
When client 3 is aggregated with 1 and 2, respectively, the test accuracy achieves 74.33\% and 82.29\%, respectively.
This phenomenon suggests that a selection strategy favoring larger firing rate differences yields faster convergence.

\begin{table}[!h]
    \centering
    \caption{Example of MNIST about the relationship between firing rate and distribution balance on MNIST.}
    \label{tab: toy experiment}
    \renewcommand\arraystretch{1.1}  
    \resizebox{1.0\linewidth}{!}{  
        \begin{tabular}{lcccc}
        \toprule
         & Client 1 & Client 2 & Client 3 & Test set \\
        \midrule
        Data distribution & 2400: 300: 300 & 300: 2400: 300 & 300: 300: 2400 & 300: 300: 300 \\
        Round 1 aggregation & \Checkmark & \Checkmark & & 59.30 \\
        Firing rate difference & 3.55 & 5.76 & 9.63 &  \\
        \multicolumn{1}{c}{\multirow{3}{*}{\begin{tabular}[c]{@{}c@{}}Round 2 aggregation \\ cases \end{tabular}}} & \Checkmark & \Checkmark & & 66.44 \\
         & \Checkmark & & \Checkmark & \textbf{74.33} \\
         & & \Checkmark & \Checkmark & \textbf{82.29} \\
        \bottomrule
        \end{tabular}
    }
\end{table}


\subsection{Client Selection Strategy}
According to the above analysis, we propose a client selection strategy based on the firing rate difference in this section.
We introduce a candidate set in our method to enhance the diversity of selected clients.
This is because this strategy will select the few clients with the most representative global distribution within a few rounds and continue to select them frequently.
In addition, a small candidate set also helps reduce the communication and power costs associated with calculating firing rate differences.
Specifically, there are three steps in each round:

\textbf{Step 1. Obtain the Candidate Client Set.} 
The server randomly samples a sub client set $\mathbb{S} (P < |\mathbb{S}| = S < N)$ as candidates.
In this way,  the diversity of clients participating in each round of aggregation can be guaranteed.

\textbf{Step 2. Calculate the Firing Rate differences.} 
In the $r^{th}$ round, for any candidate client $k \in \mathbb{S}^r$, the local SNN model is initialized with the global model $\boldsymbol{W}^r$. 
At this point, all local samples are propagated forward on the local model to count the average firing rates  $\boldsymbol{R}_k\left(\boldsymbol{W}^r\right) = \left[R_{k, 1}\left(\boldsymbol{W}^r\right), R_{k, 2}\left(\boldsymbol{W}^r\right), \cdots, R_{k, C}\left(\boldsymbol{W}^r\right)\right]^\mathbf{T}$, where $R_{k, c}\left(\boldsymbol{W}^r\right)$ is the average firing rate on the $c^{th}$ category samples, $C$ is the number of categories.
The $R_{k, c}\left(\boldsymbol{W}^r\right)$ is calculated as follows:
\begin{equation}
    R_{k, c}\left(\boldsymbol{W}^r\right) = \frac{1}{|\mathcal{D}_{k, c}|} \sum_{\xi \in \mathcal{D}_{k, c}}\frac{1}{L}\sum_{l=1}^{L}\frac{1}{\mathcal{T}|l|}\sum_{t=1}^{\mathcal{T}}o^{l,t}(\xi; \boldsymbol{W}^r),
    \label{eq: firing rate}
\end{equation}
where $\mathcal{D}_{k, c}$ is the set of samples in client $k$ of category $c$ with size $|\mathcal{D}_{k, c}|$, $\xi \in \mathcal{D}_{k, c}$ is the sample in $\mathcal{D}_{k, c}$, $L$ is the number of SNN model layers, $|l|$ is the number of neurons in layer $l$, $\mathcal{T}$ is the time window of SNN, $o^{l,t}(\xi; \boldsymbol{W}^r)$ means output spikes of the layer $l$ at the time step $t$ with the model $\boldsymbol{W}^r$ on the sample $\xi$.

Then the local model is trained $e$ epochs on the local data, and is updated to $\boldsymbol{W}_k^{r+1}$.
With the updated model, the updated firing rates $\boldsymbol{R}_k\left(\boldsymbol{W}_k^{r+1}\right)$ can be calculated as the same as the Eq. \eqref{eq: firing rate}.

The firing rate difference $\Delta \boldsymbol{R}_k^r$ of client $k$ is defined as the sum of the average firing rate difference of the local model over each category samples. 
$\Delta \boldsymbol{R}_k^r$ is calculated as follows:
\begin{equation}
    \Delta \boldsymbol{R}_k^r = \sum_{c=1}^{C} \left(R_{k, c}\left(\boldsymbol{W}_k^{r+1}\right) - R_{k, c}\left(\boldsymbol{W}^{r}\right)\right)^2.
    \label{eq: firing rate difference}
\end{equation}
This process is illustrated in the local training part of Figure \ref{fig: SFedCA}.

\textbf{Step 3. Select Clients with High Firing Rate Difference.}
From the candidate set $\mathbb{S}^r$, the server selects $P$ clients with the highest $\Delta \boldsymbol{R}_k^r$ to construct the active client set $\mathbb{P}^r$.
The server aggregates the local models of selected clients to obtain the global model $\boldsymbol{W}^{r+1}$ at the $(r+1)^{th}$ round. 
Our method takes the FedAvg as the basic FL method, and the global model is given by averaging the local models on $\mathbb{P}$.

The overall process of SFedCA is shown in Algorithm \ref{alg: SFedCA}.

\begin{table*}[!h]
    \centering
    \caption{Accuracy comparison with different methods on different datasets.}
    \label{tab: performance verification}
    \renewcommand\arraystretch{1.1}  
    \resizebox{1.0\linewidth}{!}{  
        \begin{tabular}{l|cccc|cccc|cccc}
        \hline
        \multirow{2}{*}{Methods} & \multicolumn{4}{c|}{MNIST}  & \multicolumn{4}{c|}{Fashion-MNIST} & \multicolumn{4}{c}{CIFAR-10} \\
        \cline{2-13}
          & $Dir(0.3)$ & 2Shards & $Dir_{100}(0.3)$ & $CI(3:1; 0.3)$ & $Dir(0.3)$ & 2Shards & $Dir_{100}(0.3)$ & $CI(3:1; 0.3)$ & $Dir(0.3)$  & 2Shards & $Dir_{100}(0.3)$ & $CI(3:1; 0.3)$ \\
        \hline			
        FedAvg \cite{mcmahan2017communication} & 96.51 & 95.15 & 97.97 & 93.36 & 69.44 & 67.13 & 72.71 & 65.33 & 56.99 & 34.25 & 54.00 &  45.69 \\
        FedSNN \cite{venkatesha2021federated} & 95.80 & 93.58 & 97.34 & 77.10 & 68.30 & 64.07 & 65.89 & \textbf{62.49} & 60.32 & 26.49 & 67.50 &  53.23 \\
        FedSNN+DivFL \cite{balakrishnan2022diverse} & 85.80 & 92.49 & 97.23 & 85.30 & 62.50 & 59.45 & 64.21 & 61.01 & 58.76 & 28.67 & 62.67 & 49.36 \\
        FedSNN+FedMoS \cite{Wang2023FedMoSTC} & 95.95 & 92.64 & \textbf{98.29} & 93.69 & 67.87 & 63.02 & \textbf{73.41} & 60.46 & 58.07 & 29.20 & \textbf{69.53} & 53.91 \\
        \textbf{SFedCA (Our)} & \textbf{96.37} & \textbf{95.39} & 98.21 & \textbf{94.46} & \textbf{71.03} & \textbf{65.60} & 72.50 & 61.37 & \textbf{63.50} & \textbf{46.53} & 67.01 & \textbf{55.93}  \\
        \hline
        \end{tabular}
    }
\end{table*}

\begin{algorithm}[!h]
    \SetKwFunction{Average}{Average}
    \SetKwFunction{TrainLocal}{TrainLocal} 
    \SetKwInOut{Input}{Input}
    \SetKwInOut{Output}{Output}
    \SetKw{In}{in}
    
    \Input{number of clients $N$, number of selected clients $P$, } 
    \Output{the final global model $\boldsymbol{W}$}
    \BlankLine 

    Initialize global model $\boldsymbol{W}^0$ \;
    \For{each round $r = 0, 1, \cdots$}{
        Randomly sample the candidate clients $\mathbb{S}^r$ \;
        \For{each client $k \in \mathbb{S}^r$ in parallel}{

            Calculate the firing rate $\boldsymbol{R}_k\left(\boldsymbol{W}^r\right)$ with the $r^{th}$ global model $\boldsymbol{W}^r$ on $\mathcal{D}_k$\;
            $\boldsymbol{W}_k^r \leftarrow \boldsymbol{W}^r$ \;
            $\boldsymbol{W}_k^{r+1} \leftarrow$ \TrainLocal{$\mathcal{D}_k; \boldsymbol{W}_k^r$} \;
            Calculate the firing rate $\boldsymbol{R}_k\left(\boldsymbol{W}_k^{r+1}\right)$ with the $(r+1)^{th}$ local model $\boldsymbol{W}_k^{r+1}$ on $\mathcal{D}_k$\;
            $\Delta \boldsymbol{R}_k^r \leftarrow \sum_{c=1}^{C} \left(R_{k, c}\left(\boldsymbol{W}_k^{r+1}\right) - R_{k, c}\left(\boldsymbol{W}^{r}\right)\right)^2$ \;
        }
        Construct the active client set $\mathbb{P}^r$ using the $p$ clients with the highest firing rate difference $\Delta \boldsymbol{R}_k^r$ \;
        $\boldsymbol{W}^{r+1} \leftarrow $ \Average{$\{(\boldsymbol{W}_k^{r+1}\}_{k \in \mathbb{P}^r}$} \;
        $\boldsymbol{W} \leftarrow \boldsymbol{W}^{r+1}$ \;
    }
    \KwRet{final model $\boldsymbol{W}$}
    \caption{SFedCA}
    \label{alg: SFedCA} 
\end{algorithm}

\section{Experiments}

\subsection{Experimental Setups}
\subsubsection{Models and Datasets}
Experiments are conducted on three benchmark datasets: MNIST \cite{lecun1998gradient}, Fashion-MNIST \cite{xiao2017fashion} and CIFAR-10 \cite{krizhevsky2009learning}.
On MNIST, we use a simple structure with 2 convolutional (Conv) layers and 2 fully-connected (FC) layers.
The VGG-5 structure is implemented on Fashion-MNIST.
On CIFAR-10, we take the more complex AlexNet as the backbone.

We take four methods to simulate the data heterogeneity:
\begin{itemize}
    \item \textbf{$\bm{Dir (\alpha)}$}: following \cite{hsu2019measuring}, we use the Dirichlet distribution to construct data partitions among all the clients, where the sample scales and categories are both unbalanced. The parameter $\alpha$ determines the degree of data heterogeneity between clients. 
    \item \textbf{$\bm{Dir_N(\alpha)}$}: following \cite{wang2021novel}, all the $N$ clients have all the sample categories but the sample scales obey Dirichlet distribution.
    \item \textbf{$\bm{2Shards}$}: following \cite{mcmahan2017communication}, we divide the data samples with the same label into subsets and assign 2 subsets with different labels to each client.
    \item \textbf{$\bm{CI(n_1:n_2; \alpha)}$}: following \cite{pmlr-v202-zhang23y}, we set an class-imbalanced (CI) distribution. For a 10-classes data set, the global dataset has the same amount of $n1$ data samples for five classes and $n2$ data samples for the other five classes. The sample size of all clients follows a $Dir(\alpha)$ distribution.
\end{itemize}


\subsubsection{Implementation Details}
We set the number of clients $N=100$, the number of candidates $S=10$ and the number of selected clients $P=2$.
For each client, we set the batch size as $128$, and train the local model by SGD optimizer learning rates $\eta = 0.1$, $0.003$, and $0.05$ for the MNIST, Fashion-MNIST, and CIFAR-10 datasets, respectively.

For both the MNIST and Fashion-MNIST datasets, the local model is trained for 5 epochs on each client, and for the CIFAR-10 dataset, the local model is trained for 10 epochs per client. 
The total federated learning round is set to 300.

For the SNN model, we encode the pixel values into spike trains of length $\mathcal{T} = 12$ using a spiking Conv layer \cite{zhan2024two}.
The firing threshold $u_{\theta}$ is 1 and the reset potential $u_{\text{reset}}$ is 0.

The source code of our method is shared in the supplementary material.

\subsubsection{Baselines}
We take one of the widely used spiking federated learning methods \textbf{FedSNN} \cite{venkatesha2021federated} as the baseline, which is based on the FedAvg implementation.
Based on FedSNN, we compare with the latest active client selection strategies in traditional FL, including the \textbf{DivFL} \cite{balakrishnan2022diverse} and \textbf{FedMoS} \cite{Wang2023FedMoSTC}.
\textbf{DivFL} approximates the aggregation of gradients of all the clients with that of selected clients.
\textbf{FedMoS} keeps customized momentum buffers on both server and clients tracking global and local update directions to alleviate the model discrepancy.
The client selection strategies are applied to the FedSNN in our experiments and are denoted by FedSNN+DivFL and FedSNN+FedMoS, respectively.



\begin{table*}[!h]
    \centering
    \caption{The number of communication rounds to achieve target test accuracies of different methods.$^*$}
    \label{tab: convergence round}
    \renewcommand\arraystretch{1.1}  
    \resizebox{1.0\linewidth}{!}{  
        \begin{tabular}{l|cccc|cccc|cccc}
        \hline
        \multirow{2}{*}{Methods} & \multicolumn{4}{c|}{MNIST}  & \multicolumn{4}{c|}{Fashion-MNIST} & \multicolumn{4}{c}{CIFAR-10} \\
        \cline{2-13}
          & {\makecell[c]{$Dir(0.3)$ \\ Acc 90\%}} & {\makecell[c]{2Shards \\ Acc 90\%}} & {\makecell[c]{$Dir_{100}(0.3)$ \\ Acc 95\%}} & {\makecell[c]{$CI(3:1; 0.3)$ \\ Acc 90\%}} & {\makecell[c]{$Dir(0.3)$ \\ Acc 65\%}} & {\makecell[c]{2Shards \\ Acc 60\%}} & {\makecell[c]{$Dir_{100}(0.3)$ \\ Acc 65\%}} & {\makecell[c]{$CI(3:1; 0.3)$ \\ Acc 60\%}} & {\makecell[c]{$Dir(0.3)$ \\ Acc 50\%}}  & {\makecell[c]{2Shards \\ Acc 40\%}} & {\makecell[c]{$Dir_{100}(0.3)$ \\ Acc 60\%}} & {\makecell[c]{$CI(3:1; 0.3)$\\ Acc 50\%}} \\
        \hline			
        FedAvg \cite{mcmahan2017communication} & 114 & 135 & 55 & 161 & 52 & 38 & 22 & 24 & 253 & - & - & -  \\
        FedSNN \cite{venkatesha2021federated} & 99 & 196 & 82 & - & 239 & 218 & 244 & 256 & 186 & - & 61 &  192 \\
        FedSNN+DivFL \cite{balakrishnan2022diverse} & - & 206 & 87 & - & - & - & - & 189 & 209 & - & 161 & - \\
        FedSNN+FedMoS \cite{Wang2023FedMoSTC} & 91 & 186 & 30 & 184 & 162 & \textbf{184} & \textbf{84} & \textbf{114} & 245 & - & 41 & 216 \\
        \textbf{SFedCA (Our)} & \textbf{55} & \textbf{140} & \textbf{25} & \textbf{119} & \textbf{151} & 189 & 118 & 123 & \textbf{82} & \textbf{151} & \textbf{29} & \textbf{128}  \\
        \hline
        \end{tabular}
    }
    \flushleft{
    \footnotesize{
        $^*$ - means this method failed to achieve the target accuracy within the preset number of rounds.
        }
    }
\end{table*}

\begin{figure*}[htbp]
    \centering
    \begin{subfigure}[b]{0.33\textwidth}
        \centering
        \includegraphics[width=\textwidth]{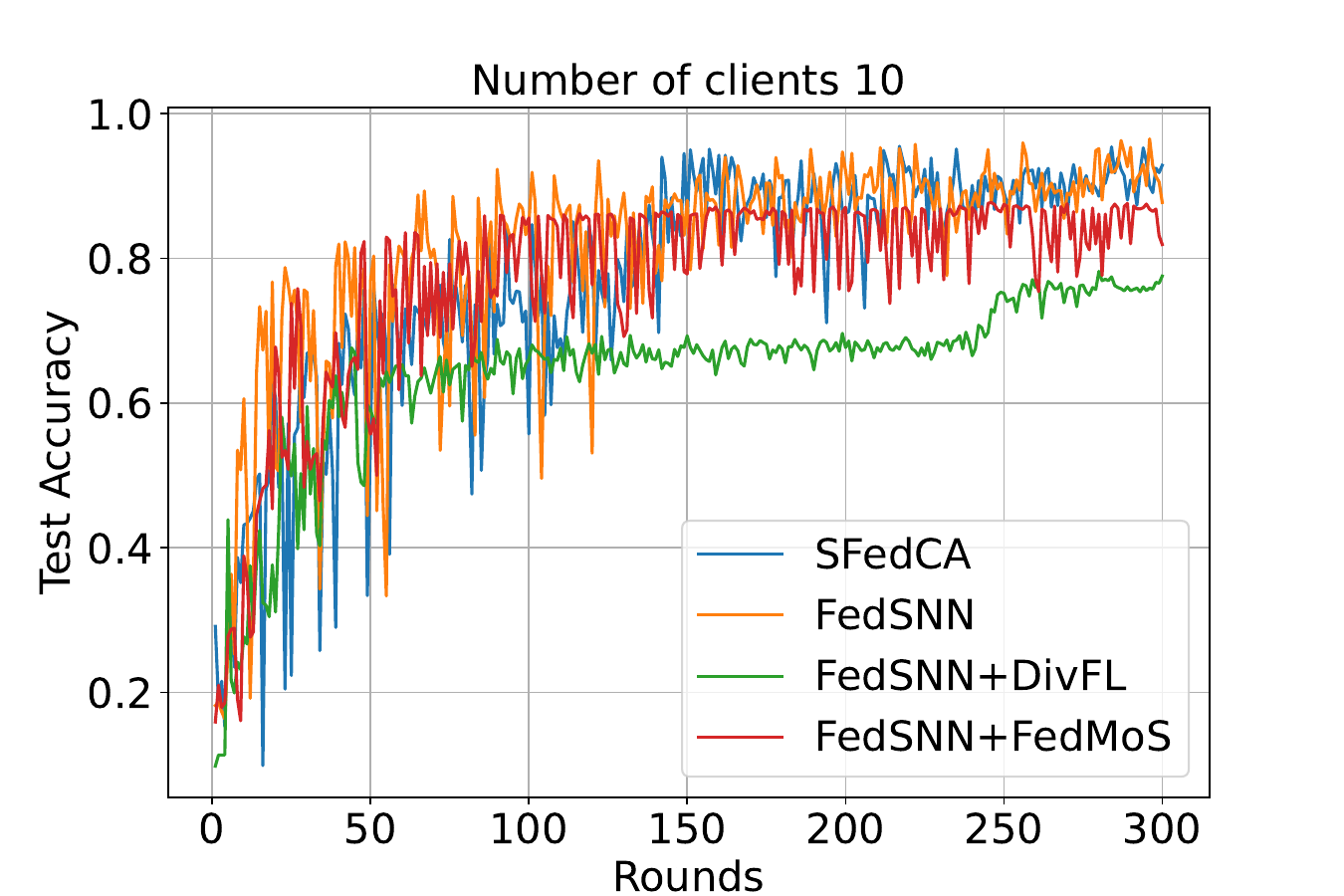}
        \caption{$N=10$}
        \label{fig: N=10}
    \end{subfigure}
    \begin{subfigure}[b]{0.33\textwidth}
        \centering
        \includegraphics[width=\textwidth]{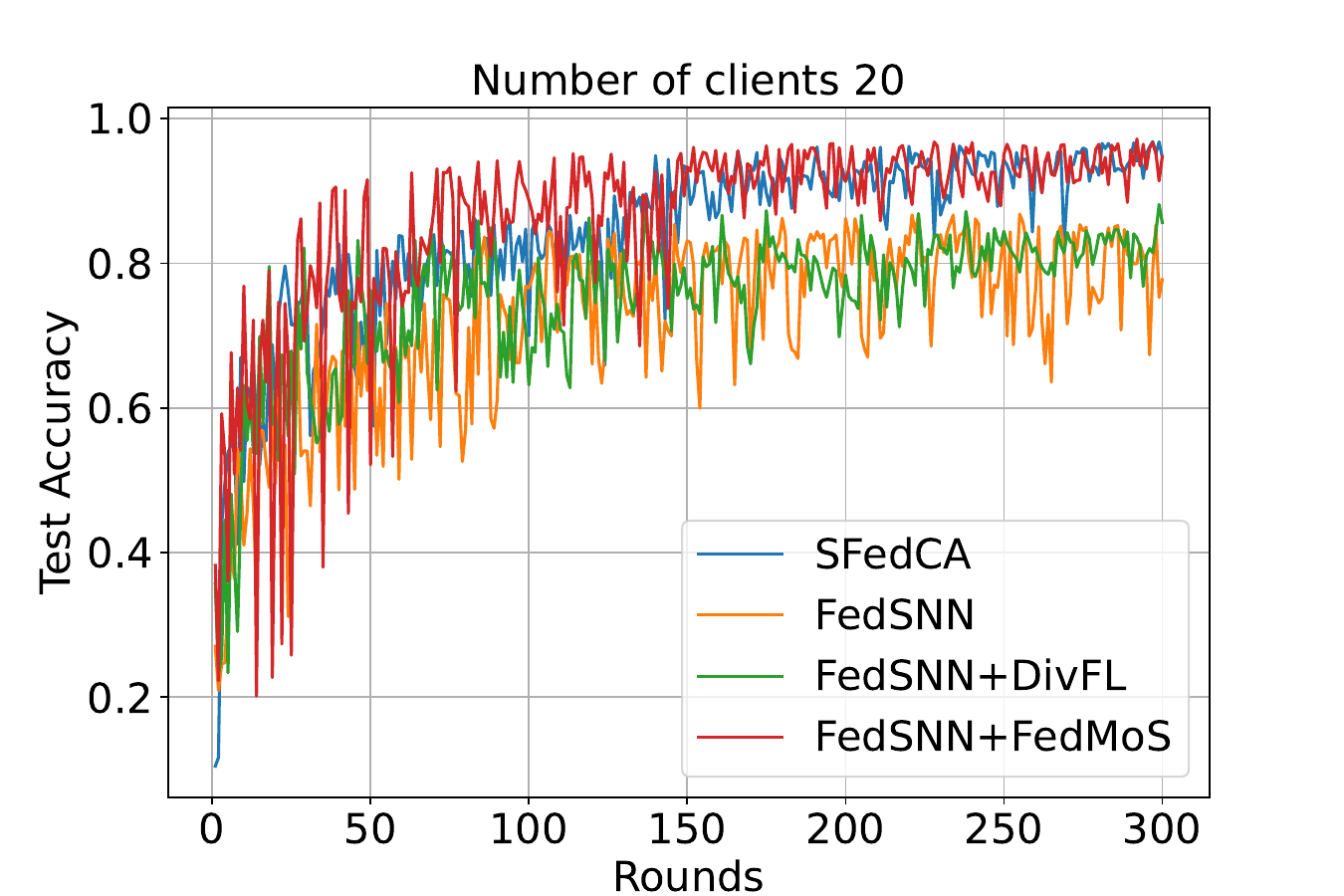}
        \caption{$N=20$}
        \label{fig: N=20}
    \end{subfigure}
    \begin{subfigure}[b]{0.33\textwidth}
        \centering
        \includegraphics[width=\textwidth]{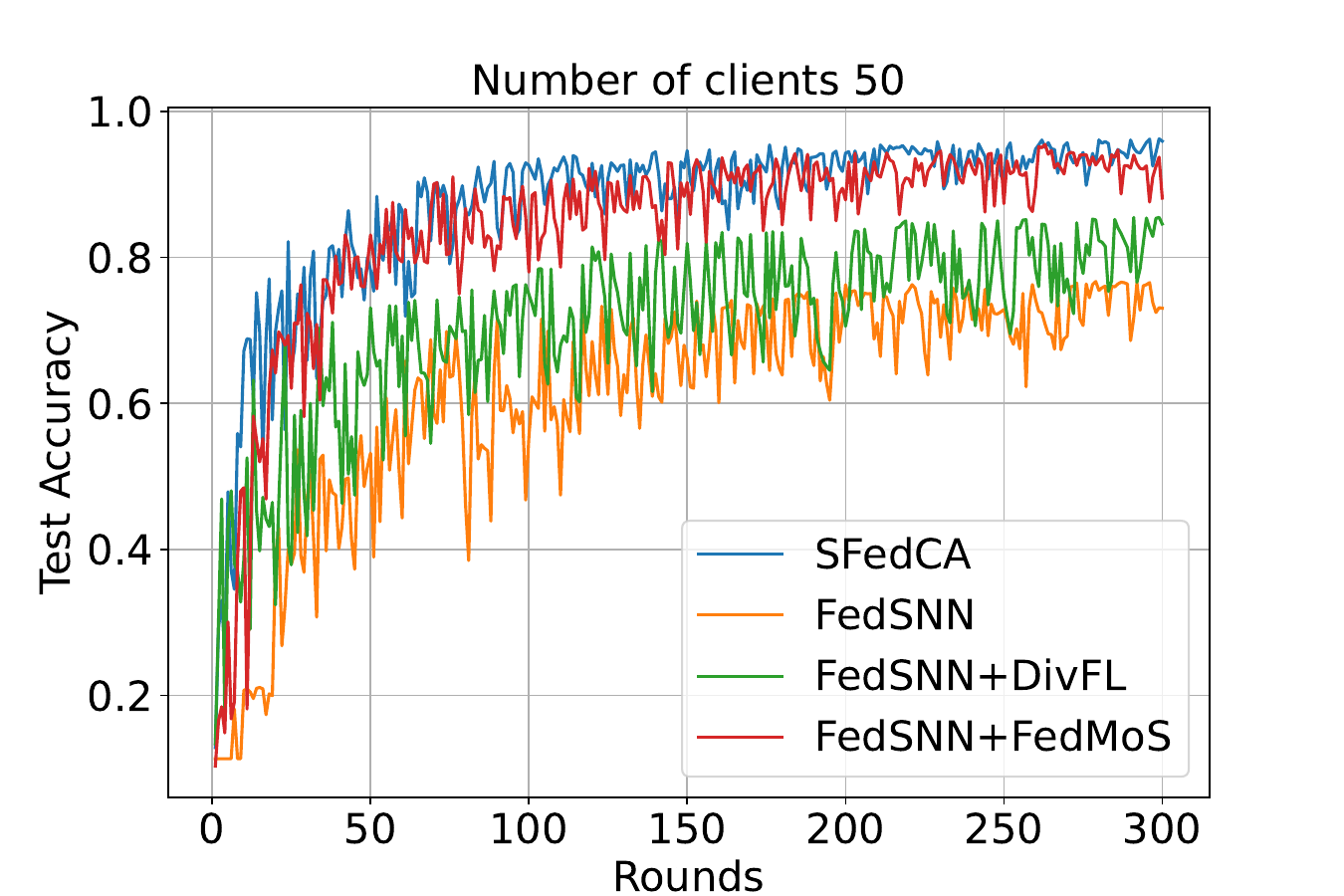}
        \caption{$N=50$}
        \label{fig: N=50}
    \end{subfigure}
    \begin{subfigure}[b]{0.33\textwidth}
        \centering
        \includegraphics[width=\textwidth]{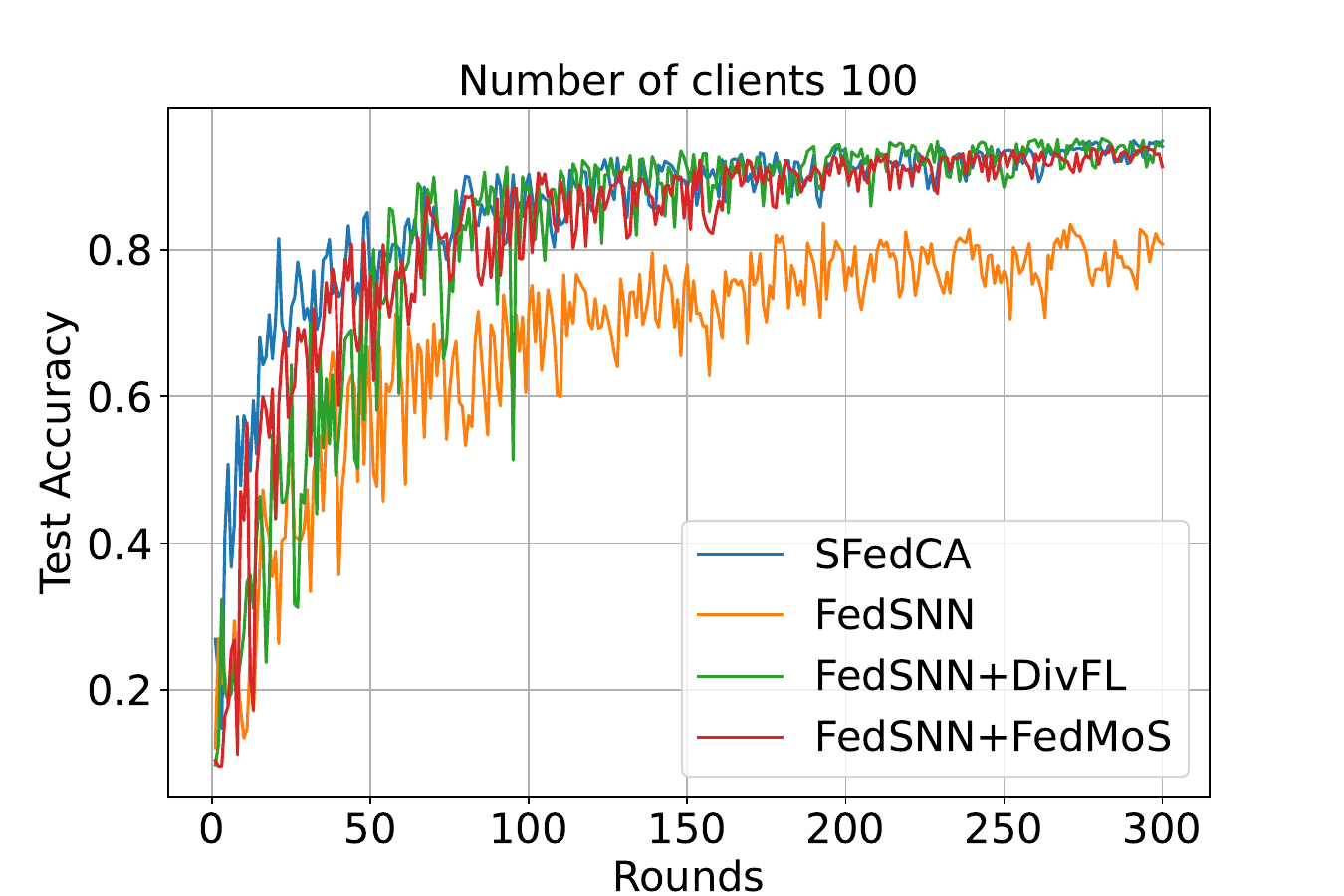}
        \caption{$N=100$}
        \label{fig: N=100}
    \end{subfigure}
    \begin{subfigure}[b]{0.33\textwidth}
        \centering
        \includegraphics[width=\textwidth]{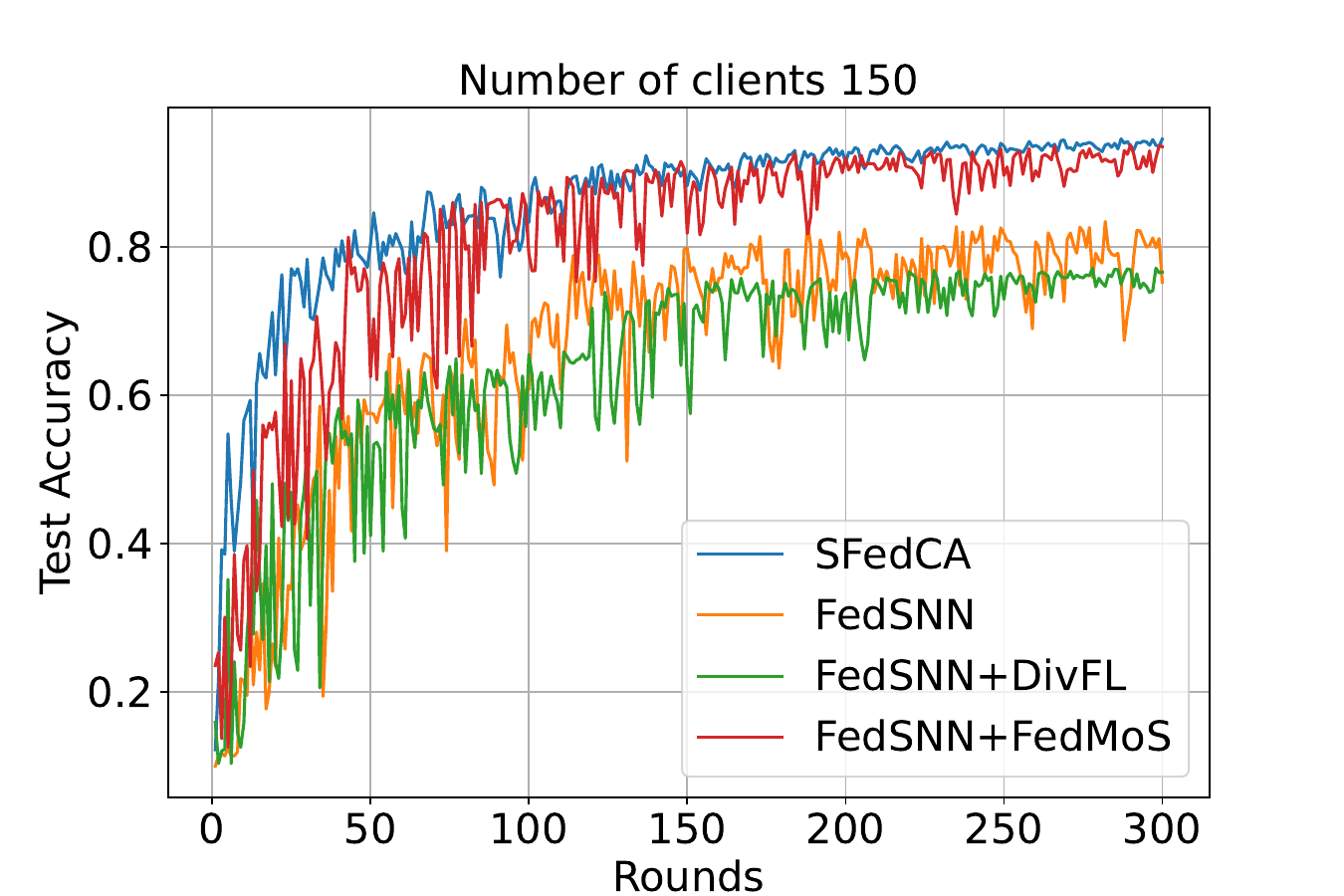}
        \caption{$N=150$}
        \label{fig: N=150}
    \end{subfigure}
    \begin{subfigure}[b]{0.33\textwidth}
        \centering
        \includegraphics[width=\textwidth]{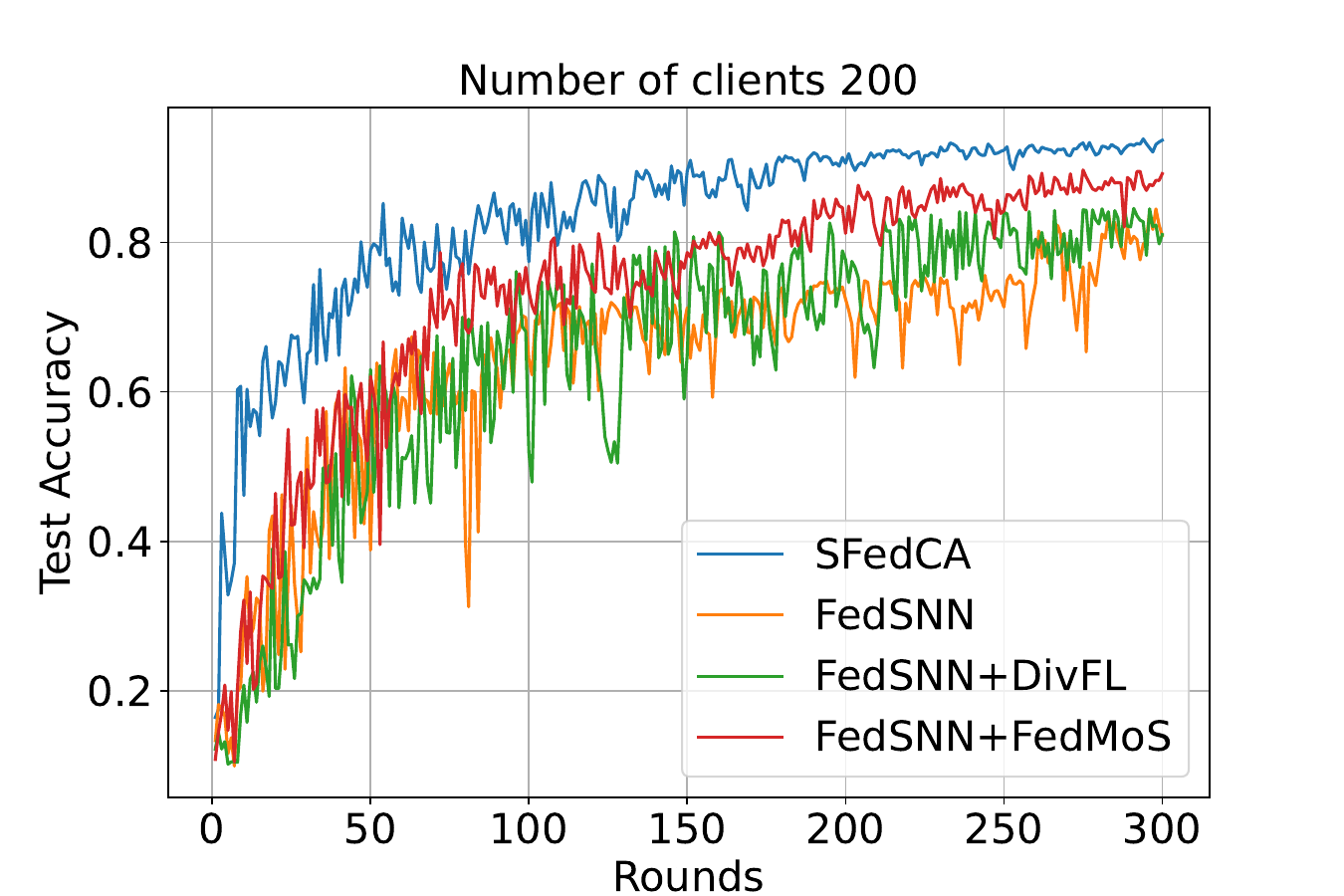}
        \caption{$N=200$}
        \label{fig: N=200}
    \end{subfigure}
    \caption{The convergence curves of SFedCA and other spiking FL methods on MNIST with different client numbers.}
    \label{fig: different client numbers}
\end{figure*}

\subsection{Comparison Results}
In this section, we conduct experiments on MNIST, Fashion-MNIST and CIFAR-10 with four different non-IID types.
To evaluate the performance of SFedCA, we compare it with the ANN-based FedAvg, FedSNN and two client selection strategies.

Table \ref{tab: performance verification} shows the comparison results of the proposed SFedCA method and other federated learning methods, in which the highest accuracies of spiking FL methods are marked in bold.
SFedCA achieves the highest accuracy among spiking FL methods on most scenarios in all three datasets.
Especially on the $Dir(0.3)$ distribution of Fashion-MNIST and CIFAR10, SFedCA have the highest accuracies 71.03\% and 63.50\%, respectively.
It is worth noting that SFedCA also outperforms the ANN-based FedAvg approach in many scenarios, which have been considered a challenge in previous studies of SNNs \cite{liu2023human}.
This phenomenon is particularly evident on CIFAR-10, and even most of the spiking FL methods achieve higher accuracy than the ANN-based FL methods.
This suggests that SNN models are more advantageous than ANN models in federated learning tasks with complex samples.

The baseline FedSNN method, which does not perform active client selection, exhibits widely varying results on different distributions of different datasets, such as the lowest accuracy of 77.10\% on the $CI(3:1; 0.3)$ distribution of MNIST and the highest accuracy of 67.50\% on $Dir_{100}(0.3)$ of CIFAR-10.
This suggests that the FL method based on stochastic aggregation is highly sensitive to the sample distribution change and is not a stable strategy.

Although FedSNN+DivFL and FedSNN+FedMoS introduce active client selection strategies and outperform the vanilla FedSNN on partial scenarios, they fail to maintain higher accuracy under all distributions, such as the 2Shards of Fashion-MNIST and $Dir(0.3)$ of CIFAR-10.
This is because these methods select clients based on gradient or momentum information, whereas the gradient of the SNN model is obtained by temporal accumulation. 
This renders the SNN gradient less effective in accurately reflecting the true sample distribution information across clients. 
Consequently, such methods hard to pinpoint those clients that would optimally contribute to broadening the scope of the global sample distribution.

In general, our proposed SFedCA method achieves higher and more stable performance than the randomized and traditional active selection methods, which is attributed to the client credit assignment strategy based on the characteristics of the SNN model.

\subsection{Communications Required for Target Accuracy}
In this section, we compare the communication efficiencies of different methods mentioned in Table \ref{tab: performance verification}.

Table \ref{tab: convergence round} shows the required communication rounds to achieve target test accuracies of different methods, in which the lowest numbers in spiking FL methods are marked in bold.
The target test accuracy is set according to the final test accuracy in Table \ref{tab: performance verification}.
SFedCA requires a minimum number of communication rounds to achieve target accuracies in almost all scenarios.
On the $Dir(0.3)$ distribution of CIFAR-10, SFedCA only takes 82 rounds to achieve the target 50\% accuracy which is 104 fewer rounds than the second fastest FedSNN method.

Comparison with FedSNN shows that the client selection strategy of SFedCA has a significant improvement in the number of communication rounds.
As contrast, there are large fluctuations in the effectiveness of the other two active FL methods.
On the 2Shards of MNIST, the number of FedSNN+DivFL communication rounds is 10 more than FedSNN; 
On the $Dir(0.3)$ of CIFAR-10, FedSNN+FedMoS takes 59 more rounds than FedSNN.

Based on the above experimental results, our method achieves an improvement over the random sampling method on different datasets, and at the same time has stronger stability compared to traditional active client sampling strategies.


\subsection{Impact of Number of Participating Clients}
In this section, we validate the adaptability of our approach to different client numbers and proportions of selected clients.
Two sets of experiments are conducted on MNIST.

\textbf{Different total client numbers.}
In the first set of experiments, we fix the proportion of selected clients to be 0.1 and the candidate ratio to be 0.2, and set the total numbers of clients to be 10, 20, 50, 100, 150 and 200.
The test non-IID type is set to $CI(3:1;0.3)$.
In this setting, we compare our method with FedSNN and its variants of active forms to verify the efficiency of SFedCA.

Figure \ref{fig: different client numbers} shows the convergence curves of different methods.
SFedCA achieves the highest accuracy rate for different numbers of clients.
As the number of clients increases to 200, the gap between the other methods and our SFedCA becomes more pronounced.
When the number of clients is small (10 and 20), the accuracy differences between these methods are smaller, because the candidate set is smaller and the range of client selection is smaller.
FedSNN+FedMoS stays competitive when the number of clients is small, but shows more fluctuations when $N$ is increased to 150, and lags significantly behind SFedCA when $N$ is 200.

\textbf{Different selected client numbers.}
In the second set of experiments, we fix the total number of clients to 100 and the number of candidates to 20.
The numbers of selected clients are 2, 5, 10, 15 and 20.
The test non-IID type is set to $Dir(0.3)$.
In this setting, we take the SNN-based FedAvg which aggregates all the clients as the baseline to explore the client efficiency of SFedCA.

\begin{figure}[h]
  \centering
  \includegraphics[width=0.6\linewidth]{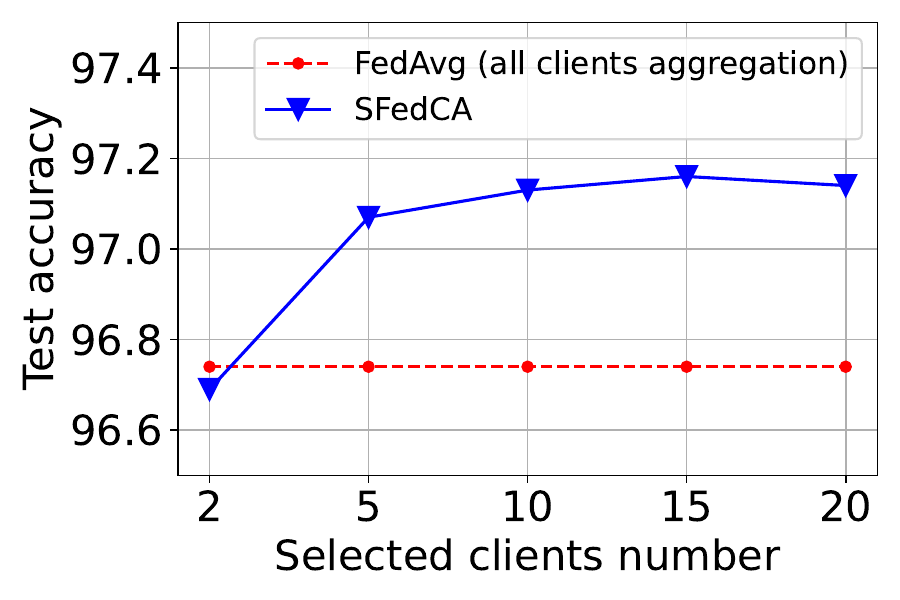}
  \caption{The performance of SFedCA with different selected client numbers.}
  \label{fig: different selected numbers}
\end{figure}

Figure \ref{fig: different selected numbers} shows the test accuracies of SFedCA with different numbers of selected clients. 
The accuracy of the FedAvg considering all the clients is 96.74\%. 
When only two clients are selected per round, SFedCA is less accurate than FedAvg in Figure 5 only 96.69\%.
As the number of selected clients increases to 5, the accuracy of SFedCA exceeds that of FedAvg.
Upon increasing the number to 20, all candidate sets participate in the aggregation, and clients with redundant information instead interfere with the optimization direction of the global model.
This will lead to a downward trend in model accuracy.
This suggests that actively selected clients provide higher training efficiency.

The experimental results in this section demonstrate that SFedCA can effectively select clients suitable for global model optimization under different total numbers of clients and numbers of participating clients.

\subsection{Robustness to Data Noise}
In this section, we analyze the robustness of the global model trained by SFedCA against noise perturbations in data.
Specifically, we add the Gaussian noise to the inputted test data following \cite{kim2023exploring}.
Gaussian noise simulates random errors often encountered in real-world scenarios, and we set its L2-norm from 0.0 to 0.5 of the L2-norm of the given input data. 

\begin{figure}[htbp]
    \centering
    \begin{subfigure}[b]{0.23\textwidth}
        \centering
        \includegraphics[width=\textwidth]{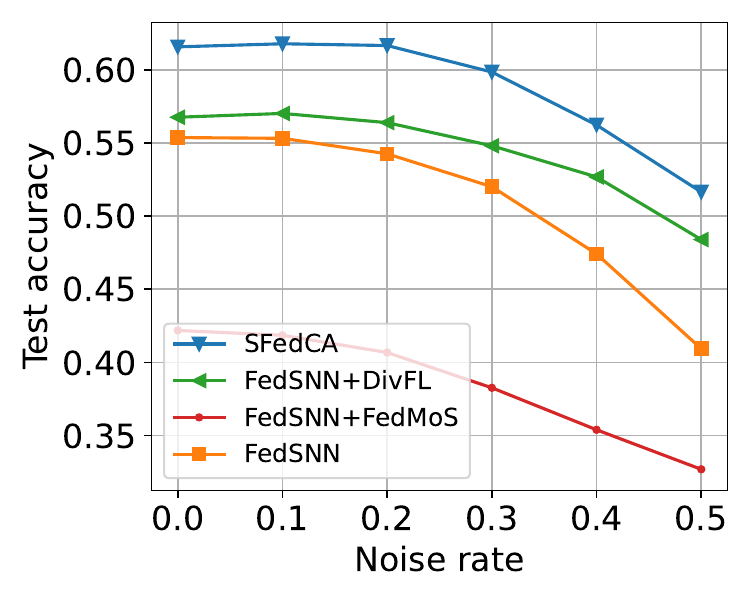}
        \caption{$Dir(0.3)$}
        \label{fig: gaussian result dir_0.3}
    \end{subfigure}
    \begin{subfigure}[b]{0.23\textwidth}
        \centering
        \includegraphics[width=\textwidth]{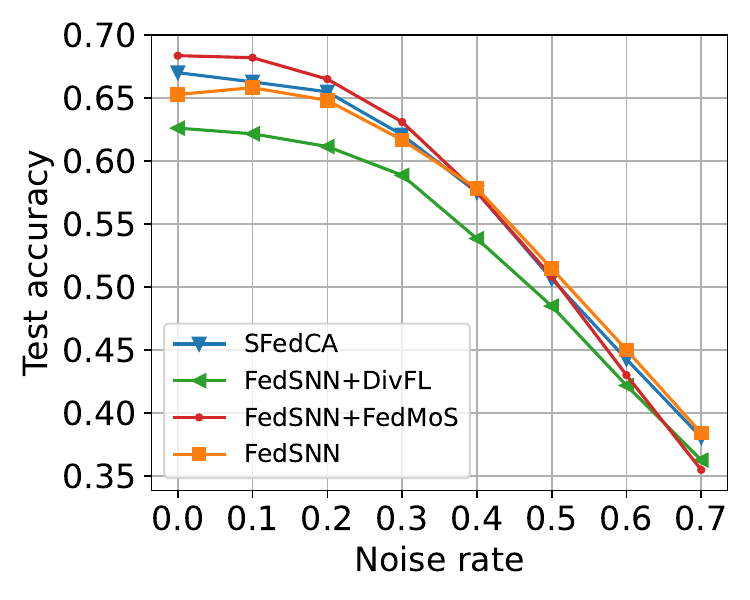}
        \caption{$Dir_{100}(0.3)$}
        \label{fig: gaussian result dir_100_0.3}
    \end{subfigure}
    \caption{Accuracy of different methods on the noised CIFAR-10 data.}
    \label{fig: gaussian result}
\end{figure}

Figure \ref{fig: gaussian result} shows the test accuracies change with different noise rates.
Figure \ref{fig: gaussian result dir_0.3} records the test results of models trained on $Dir(0.3)$ of CIFAR-10.
The accuracy of SFedCA do not decrease until the noise rate is less than 0.3. 
When the noise rate is greater than 0.3, SFedCA and FedSNN have similar decreasing trends.
FedSNN+DivFL has a flatter downward curve and is consistently higher than FedSNN but lower than SFedCA.
Although FedSNN+FedMoS also has a flatter trend, the accuracy is not competitive.
Figure \ref{fig: gaussian result dir_100_0.3} records the test results of on $Dir_{100}(0.3)$ of CIFAR-10.
Although the FedSNN+FedMoS method has high initial accuracy,  it decreases most rapidly as the noise rate increases.
Our SFedCA method maintains a similar decrease rate as FedSNN for both data distributions.
Compared to other active methods, the robustness of SFedCA to noise is not degraded by the loss of client randomness.

\subsection{Client Data Distribution Analysis}
\begin{figure}[h]
  \centering
  \includegraphics[width=\linewidth]{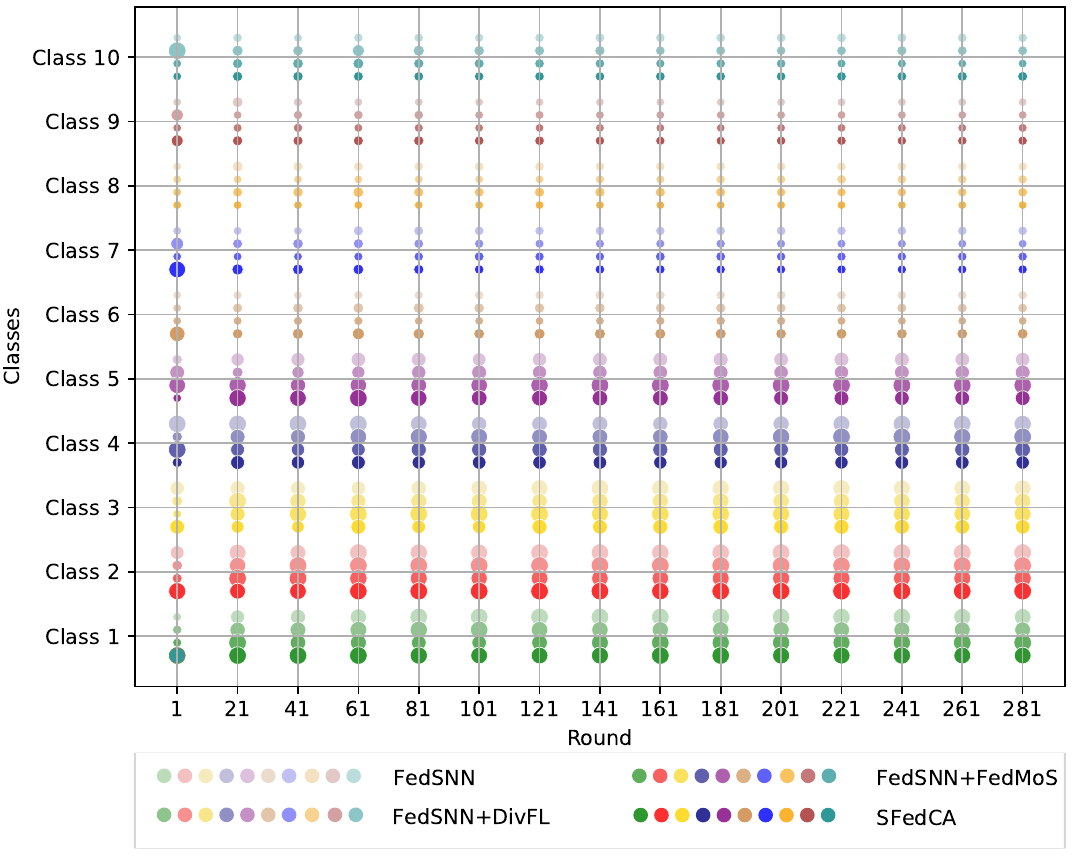}
  \caption{Proportion of sample distribution of clients selected by different methods on the $CI(3:1;0.3)$ of CIFAR-10. Larger points indicate a larger percentage.}
  \label{fig: distribution}
\end{figure}

To analyze the effect of our method on the balance of the data distribution, we visualize the data distribution of selected clients in this section.
We test methods on the $CI(3:1;0.3)$ distribution of CIFAR-10 which have the largest difference in category proportions.
Figure \ref{fig: distribution} shows the variation of the cumulative proportions of the selected client samples with training rounds.

Since the first five categories in this distribution are three times as numerous as the last five, the distribution of client samples randomly sampled by FedSNN obeys this ratio.
The proposed SFedCA has a more balanced distribution, i.e., the proportion of classes 1-5 samples in round 281 is the smallest of all methods, and classes 6- 10 also have a higher share. 
This further suggests that SFedCA is effective in balancing the distribution of samples indirectly used by the global model.

\subsection{Comparison of Energy Consumption }
To demonstrate the low power consumption advantage of SFedCA, we calculate the theoretical power cost of our method and ANN-based FedAvg.
Following \cite{zhou2023spikformer}, we calculate the number of operations (OPs) in models. 
For the ANN model, OPs is the number of floating point operations (FLOPs).
For SNN, OPs is the number of synaptic operations (SOPs), which is formulated as follows:

\begin{equation}
    \mathrm{SOPs}(\mathcal{D}_t) = \frac{1}{|\mathcal{D}_t|} \boldsymbol{R}_{\mathcal{D}_t}\left(\boldsymbol{W}\right) \times \mathcal{T} \times \mathrm{FLOPs}(\mathcal{D}_t),
\end{equation}
where $\mathcal{D}_t$ is the test dataset; 
$\boldsymbol{R}_{\mathcal{D}_t}\left(\boldsymbol{W}\right)$ represents the firing rate of the SNN model on $\mathcal{D}_t$ with parameters $\boldsymbol{W}$; 
$\mathrm{FLOPs}(\mathcal{D}_t)$ denotes the number of floating point operations, i.e., MAC (multiply and accumulate) operations;
$\mathrm{SOPs}(\mathcal{D}_t)$ represents the number of Accumulate (AC) operations based on spikes. 
Take the implementation on 45 nm hardware as an example, one FLOPs and one SOPs computation
consume 4.6 pJ and 0.9 pJ of energy, respectively.

\begin{figure}[htbp]
    \centering
    \begin{subfigure}[b]{0.23\textwidth}
        \centering
        \includegraphics[width=\textwidth]{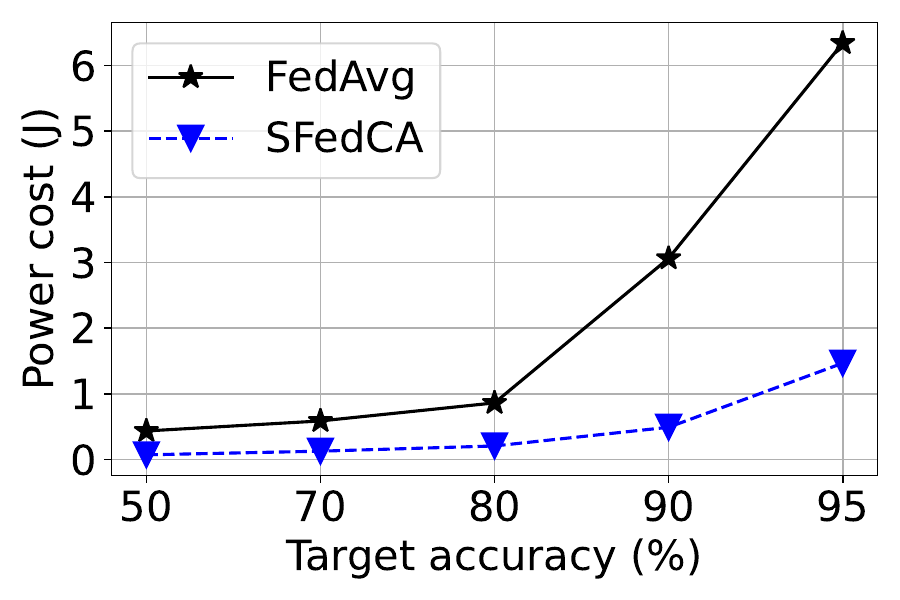}
        \caption{$Dir(0.3)$}
        \label{fig: training power dir(0.3)}
    \end{subfigure}
    \begin{subfigure}[b]{0.23\textwidth}
        \centering
        \includegraphics[width=\textwidth]{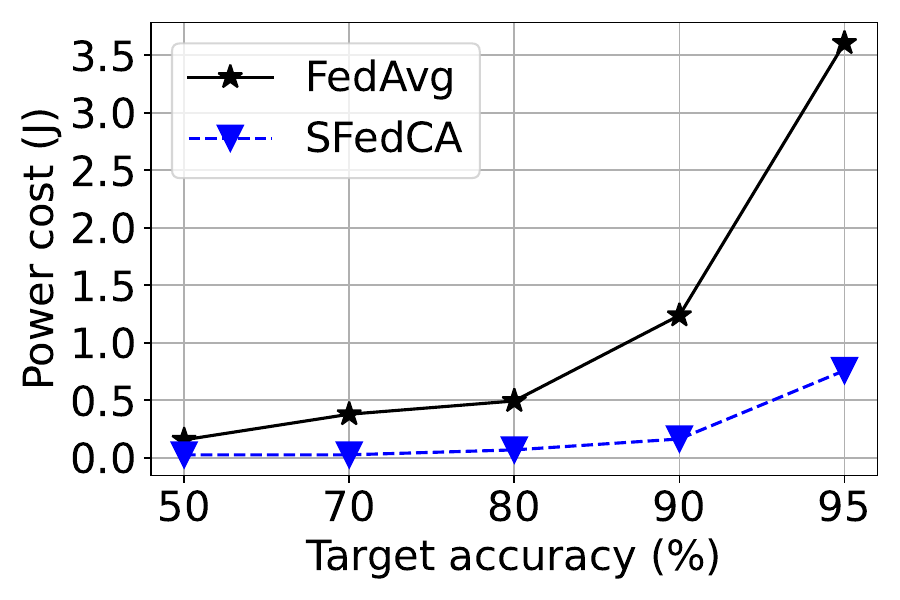}
        \caption{$Dir_{100}(0.3)$}
        \label{fig: training power dir_100(0.3)}
    \end{subfigure}
    \caption{Training power cost with different target accuracies on MNIST.}
    \label{fig: training power}
\end{figure}

We conduct experiments on the the $Dir(0.3)$ and $Dir_{100}(0.3)$ of MNIST, using the same network architecture for both ANN-based FedAvg and SFedCA. 
Figure \ref{fig: training power} shows the training power cost changes of these two methods to achieve the target test accuracies.
SFedCA keeps an extremely low consumption power cost at all the target accuracies.
In addition, the increase in training power consumption of SFedCA consistently grows slower as the target accuracy increases from 50\% to 90\%.
The training power growth of FedAvg ramps up as it reaches 90\% accuracy, and at 95\%, the cost required is close to five times that of SFedCA.

\begin{table}[!h]
    \centering
    \caption{Average inference power costs ($\mu J$) of different methods.}
    \label{tab: test power}
    \renewcommand\arraystretch{1.1}  
    \resizebox{0.8\linewidth}{!}{  
        \begin{tabular}{lcccc}
        \toprule
        \multirow{2}{*}{Methods} & \multicolumn{2}{c}{MNIST} & \multicolumn{2}{c}{CIFAR-10} \\
        \cmidrule{2-5}
         & $Dir(0.3)$ & $Dir_{100}(0.3)$ & $Dir(0.3)$ & $Dir_{100}(0.3)$ \\
        \midrule
        FedAvg & 13.478 & 13.478 & 386.308 & 386.308 \\
        \textbf{SFedCA (Our)} & \textbf{6.858} & \textbf{6.809} & \textbf{51.558} & \textbf{57.858} \\
        \bottomrule
        \end{tabular}
    }
\end{table}

Table \ref{tab: test power} shows the average inference power costs of FedAvg and SFedCA on a single sample of MNIST and CIFAR-10. 
On the simple MINST dataset, SFedCA is half the cost of FedAvg; on the more complex CIFAR-10, SFedCA saves 7 times the inference power.

The above experimental analysis demonstrates the significant advantages of SFedCA over the traditional ANN-based FL approach in terms of training and inference power consumption.

\section{Conclusion}
In this paper, to address the neglect of statistical heterogeneity in existing spiking FL methods, we propose an SFedCA method with an active client selection strategy.
This approach introduces the concept of credit alignment from SNNs, and uses the difference in firing rates before and after local model training to compute the client credit.
By selecting clients with high credit, the global model can indirectly reach a more balanced data distribution, thus speeding up model convergence and improving accuracy.
We conducted experiments on four different non-IIDs in three datasets.
The experimental results indicate that SFedCA can effectively improve the accuracy of the global model and reduce the communication rounds.
SFedCA is more stable applied to different datasets and distribution scenarios than traditional active selection strategies.
It has a significant power advantage over the ANN-based FL method, with inference power savings of up to 7 times.
We note that although SFedCA helps to improve the sample balance of the global model, it is still not optimal. 
Therefore, in our future work, we will further investigate more accurate measurements of client credits to improve the global model performance.

\begin{acks}
To Robert, for the bagels and explaining CMYK and color spaces.
\end{acks}

\bibliographystyle{ACM-Reference-Format}
\bibliography{mybib}










\end{document}